# Reformulating Inference Problems Through Selective Conditioning

**Paul Dagum and Eric Horvitz**
Section on Medical Informatics
Stanford University School of Medicine
and
Palo Alto Laboratory
Rockwell International Science Center
444 High Street
Palo Alto, California 94301

## Abstract

We describe how we selectively reformulate portions of a belief network that pose difficulties for solution with a stochastic-simulation algorithm. With employ the *selective conditioning* approach to target specific nodes in a belief network for decomposition, based on the contribution the nodes make to the tractability of stochastic simulation. We review previous work on BNRAS algorithms—randomized approximation algorithms for probabilistic inference. We show how selective conditioning can be employed to reformulate a single BNRAS problem into multiple tractable BNRAS simulation problems. We discuss how we can use another simulation algorithm—logic sampling—to solve a component of the inference problem that provides a means for knitting the solutions of individual subproblems into a final result. Finally, we analyze tradeoffs among the computational subtasks associated with the selective-conditioning approach to reformulation.

## 1 INTRODUCTION

We have developed a method for identifying and reformulating variables in a belief network to maximize the efficiency of probablistic inference with a stochastic-simulation algorithm. The approach is based on the selection of nodes for decomposition through conditioning by considering how the decomposition will affect inference efficiency. Although we focus on simulation-based inference with belief networks, we believe our approach has application to the solution of other difficult computational problems by providing a methodology for intelligently decomposing the most difficult components of a problem instance, and for directing subproblems to the most suitable solution procedures.

We shall first describe BNRAS algorithms for probabilistic inference. These include the BNRAS algorithm by Chavez and Cooper [3, 2], $\mathcal{D}$-BNRAS by Dagum and Chavez [6], and CS-BNRAS by Dagum et al. [9]. We shall discuss how we can parameterize the runtime of these algorithms in terms of a parameter $\mathcal{D}$, a function of the dependency structure and the conditional probabilities of a belief network. We shall then show how we can decrease the maximal $\mathcal{D}$ associated with a belief network by targeting nodes of a network that contribute significantly to the value of $\mathcal{D}$ for decomposition through conditioning.

Selective conditioning decomposes complex portions of a belief network into subproblems that can be solved efficiently with BNRAS algorithms. Solving the global inference problem requires taking a weighted sum of the results of the subproblem inferences. Thus, the final solution requires a method for computing the weights on each subproblem. Although inference within subproblems is amenable to approximation with BNRAS, the algorithm cannot be used to efficiently compute the weights on subproblems. We show that the weights are ideally suited for efficient approximation by a different simulation algorithm: we apply a modification of Henrion's logic sampling [15] for this task. Our modification of logic sampling employs a Dirichlet stopping rule [8] that allows us to generate the needed probability distribution over conditioned nodes in optimal time.

## 2 RELATED WORK

The method of conditioning was introduced by Pearl for decomposing multiply connected belief networks into a set of singly connected belief network problems through identification of the cutset [20]. In our work, we do not seek to identify and instantiate a cutset to completely decompose a multiply connected network. Rather we employ selective conditioning to identify portions of a belief network which pose the most difficult problems to solution with a simulation algorithm. Through selective conditioning, we compose a reformulation search space of subsets of multiply connected nodes, and seek to choose a set of nodes to decompose the network most effectively.

In related work on reformulation, Cooper and Chin have examined the reformulation of a belief network through Bayesian arc-reversal in an attempt to eradicate small-valued conditional probabilities in the network [4]. Breese and Horvitz examined the ideal trade-off in reformulation versus solution-execution effort in searching for cutsets for application of the method of conditioning [1]. In other work, Horvitz posed the use of selective conditioning as a means for *topologically editing* belief network-problem instances into sets of singly and multiply connected belief-network subproblems for analysis by combinations of algorithms, each best suited to the alternative subproblems [16]. Horvitz et al. employed the principles of cutset conditioning to develop a flexible inference algorithm that allows for varying amounts of incompleteness in conditioning [17]. Suermondt et al. describe the value of combining conditioning with the clique-tree methodology of Lauritzen and Spiegelhalter [18] for solving problems with special nodes (e.g., disease nodes in medical belief networks) that play a role of primary causation, as ancestor to almost all other nodes in the belief network [22].

## 3 RANDOMIZED APPROXIMATION SCHEMES

We shall use $B$ to denote a binary-valued belief network on $n$ nodes $\{X_1, ..., X_n\}$. For any node $X_i$, and parents $\mathbf{u}_{X_i}$, a belief network specifies a conditional probability function $\Pr[X_i|\mathbf{u}_{X_i}]$. The full joint probability distribution specified by a belief network can be calculated by taking the product of the conditional probabilities. Thus,

$$\Pr[X_1, ..., X_n] = \prod_{i=1}^{n} \Pr[X_i|\mathbf{u}_{X_i}].$$

Probabilistic inference in belief networks refers to the computation of $\Pr[X = x|E]$, for some set of nodes $X$ instantiated to $x$ and conditioned on evidence $E$.

Randomized approximation schemes (RAS) for probabilistic inference [3, 6] are a class of stochastic-simulation algorithms. Simulation procedures for inference estimate the value of an exact result by determining the fraction of successes of a Bernoulli process. Let $\phi$ denote the value of $\Pr[X = x|E]$. Stochastic simulation algorithms for probabilistic inference provide an estimate $\mu$ of $\phi$. Beyond randomized approximation schemes, simulation algorithms include logic sampling [15], straight simulation [19], and likelihood weighting [21, 13].

A simulation algorithm is a *randomized approximation scheme* if, on input parameters $\epsilon$ and $\delta$, the algorithm outputs an estimate $\mu$ that satisfies

$$\Pr[\phi(1+\epsilon)^{-1} \le \mu \le \phi(1+\epsilon)] > 1 - \delta \quad (1)$$

## 4 RAS ALGORITHMS FOR INFERENCE

The BNRAS algorithms, including BNRAS, $\mathcal{D}$-BNRAS, and CS-BNRAS, represent a family of algorithms that provide approximations to probabilistic inferences satisfying Equation 1. The operation of BNRAS algorithms can be decoupled into a *trial-generation phase* and a *scoring phase*. The trial-generation phase generates belief network instantiations consistent with the observed evidence. Thus, for unobserved nodes $Z$ and evidence $E$, the instantiation $Z = z$ is generated with probability $\Pr[Z = z|E]$. If we desire to approximate the inference $\Pr[X = x|E]$, the scoring phase computes the fraction of trials that produce instantiations consistent with the inference $\Pr[X = x|E]$.

Dagum and Chavez showed that the efficiency of BNRAS algorithms is independent of the inference query, but is critically dependent on the efficiency of the trial-generation phase [6]. In addition they showed that the efficiency of the trial-generation phase depends on the *dependence value*, an easily computable quantity of the belief network.

## 5 DEPENDENCE VALUE OF BELIEF NETWORKS

Dagum and Chavez [6], parameterize belief networks by their *dependence value*, $\mathcal{D}_E \ge 1$. The dependence value of a belief network depends on the evidence $E$ that has been observed. The dependence value provides a measure of the cumulative strength of the dependencies among nodes in a belief network that are encoded by the conditional probabilities associated with each node.

For each node $X_i$, we define $l_i$ and $u_i$ as the greatest and smallest numbers, respectively, such that, for instantiation $x_i$ of $X_i$, and for all instantiations of the nodes in $\mathbf{u}_{X_i}$ that are not evidence nodes,

$$l_i \le \Pr[x_i|\mathbf{u}_{X_i}] \le u_i. \quad (2)$$

It follows that

$$(1 - u_i) \le \Pr[\overline{x}_i|\mathbf{u}_{X_i}] \le (1 - l_i), \quad (3)$$

where $\overline{x}_i$ denotes $1 - x_i$. Note that $l_i > 0$ and $u_i < 1$, since we are assuming that no complete instantiation of the network has zero probability. If $X_i$ is not an evidence node, then we define $\lambda_i = \max\left(\frac{u_i}{l_i}, \frac{1-l_i}{1-u_i}\right)$. If $X_i$ is an evidence node, and $X_i = x_i$, then $\lambda_i = \frac{u_i}{l_i}$. If $X_i = \overline{x_i}$ then $\lambda_i = \frac{1-l_i}{1-u_i}$. When $X_i$ is a prior node, or when $\mathbf{u}_{X_i}$ contains only evidence nodes, then $\lambda_i = 1$.

**Definition** For a belief network $B$, the *dependence value* is given by

$$\mathcal{D}_E = \prod_{i=1}^{n} \lambda^2_i.$$

By definition, $\mathcal{D}_E \geq 1$. The trivial case where $\mathcal{D}_E = 1$ occurs when the variables representing the nodes of the belief network are all mutually independent; that is, the belief network does not contain any arcs.

## 6 DEPENDENCE VALUE AND TRACTABILITY

The time required to approximate an inference with D-BNRAS, or with CS-BNRAS, is the product of $\mathcal{D}_E^4$ and a polynomial in the number of nodes in the belief network. Thus, the dependence value is a measure of the tractability of approximation, where increases in $\mathcal{D}_E$ render approximations more intractable. The dependence value of a belief network is dominated by bounds on the conditional probabilities, given by Equations 2 and 3, that are close to 0 and 1. When the number of observed nodes $E$ increases, the bounds on the conditional probabilities move away from 0 and 1. Thus, with increasing evidence, the dependence value $\mathcal{D}_E$ decreases, and approximations that are otherwise intractable are rendered tractable.

## 7 PROBLEM REFORMULATION

We show how approximation of probabilistic inference for belief network problem instances with large dependence values can be reformulated into the approximation of a set of inference problems with small dependence values.

In the preceding section we observed that large evidence sets resulted typically in small dependence values. We achieve the greatest reduction of the dependence value when we instantiate selectively the parents of the nodes with the largest $\lambda_i$s. However, when we instantiate nodes, we change the inference that is approximated by the BNRAS algorithm. For example, suppose we wish to approximate $\Pr[X|E]$, but, because the value of $\mathcal{D}_E$ is too large for tractable approximation, we are led to instantiate nodes $\Im$. The resultant dependence value, $\mathcal{D}_{E,\Im}$, allows tractable approximation, *but* of the inference $\Pr[X|E,\Im]$. We can recover the correct inference $\Pr[X|E]$ if we observe that

$$\Pr[X|E] = \sum_{\Im} \Pr[X|E,\Im]\Pr[\Im|E], \quad (4)$$

where the summation is over all instantiations of $\Im$.

The problem reformulation by conditioning requires that we pose the inference problem $\Pr[X|E]$ as the two inference problems $\Pr[X,E]$ and $\Pr[E]$. From Bayes' Rule we can express

$$\Pr[X|E] = \frac{\Pr[X,E]}{\Pr[E]}.$$

Furthermore, a property of RAS algorithms is that the ratio of estimates that satisfy Equation 1 for $\Pr[X,E]$ and for $\Pr[E]$, is an estimate of $\frac{\Pr[X,E]}{\Pr[E]}$ that also satisfies Equation 1 — and is, therefore, an estimate of $\Pr[X|E]$. We can approximate $\Pr[X,E]$ — and similarly, $\Pr[E]$ — if we observe that

$$\Pr[X,E] = \sum_{\Im} \Pr[X,E|\Im]\Pr[\Im]. \quad (5)$$

The choice of the nodes in $\Im$ guarantees that the inferences $\Pr[X,E|\Im]$ are approximated readily using a BNRAS algorithm. However, use of Equation 5 poses two problems. First, to evaluate the sum requires us to approximate $2^{|\Im|}$ inferences, where $|\Im|$ denotes the number of nodes in $\Im$. Thus, crucial to the success of problem reformulation is the existence of small sets $\Im$ that, when instantiated, effectively reduce the dependence value. The second challenge we encounter in Equation 4 is the efficient approximation of inferences $\Pr[\Im]$. We cannot use a BNRAS algorithm because the dependence value for the case of no evidence is at least as large as $\mathcal{D}_E$, and by assumption, the size of $\mathcal{D}_E$ prohibits tractable approximations.

## 8 DIRICHLET DISTRIBUTIONS

In other work [7, 8], we exploited the conjugate relation between multinomial distributions and Dirichlet distributions to derive stopping rules for multinomial stochastic processes that appear in stochastic simulation algorithms. We give a brief review of the material in [8], and we explore how the stopping rules can be employed to generate approximations to the probability distribution $\Pr[\Im]$.

We simulate the belief network using logic sampling. The output of each trial is a complete instantiation $x_1, ..., x_n$ of the nodes generated with probability distribution $\Pr[x_1, ..., x_n]$. Let $I_i$ denote the $i$th instantiation of the nodes in $\Im$ and let $\phi_i = \Pr[I_i]$. Consider the stochastic process generated by the random variable $\zeta = \zeta(I_i)$ whose outcome must belong to one of the $K = 2^{|\Im|}$ mutually exclusive and exhaustive categories that label all possible instantiations of $\Im$. The probability that the outcome belongs to the $i$th category is given by $\phi_i$ $i = 1, ..., K$. Assume we observe $N$ outcomes of $\zeta$. Let $n_i$, $i = 1, ..., K$, denote the number of these outcomes that belong to the $i$th category. The *random vector* $\vec{n} = (n_1, ..., n_K)$ has a multinomial distribution with parameters $N$ and $\vec{\phi} = (\phi_1, ..., \phi_K)$, that is,

$$\Pr[\vec{n}|\vec{\phi},N] = \frac{N!}{n_1!\cdots n_K!}\phi_1^{n_1}\cdots\phi_K^{n_K}.$$

For a random vector with a multinomial distribution, the conjugate distribution is provided by a Dirichlet distribution. Thus, for the preceding example, let $\mu_i = \frac{n_i}{N}$, $i = 1, ..., K$, and $\vec{\mu} = (\mu_1, ..., \mu_K)$. The distribution of $\vec{\phi}$, having observed $\vec{\mu}$ and $N$, is given by the Dirichlet distribution with parameters $\vec{\mu}$ and $N$,

$$\begin{aligned}\delta(\vec{\phi}|\vec{\mu},N) &= \frac{(N-1)!}{(\mu_1 N - 1)!\cdots(\mu_K N - 1)!}\\&\quad\times\phi_1^{\mu_1 N - 1}\cdots\phi_K^{\mu_K N - 1}.\end{aligned} \quad (6)$$

For $i = 1, ..., K$, the Dirichlet distribution mean of $\phi_i$ is $\mu_i$, and the variance of $\phi_i$ is,

$$v_\delta(\phi_i) = \frac{1}{N+1}\mu_i(1-\mu_i).$$

The Dirichlet distribution tells us how $\vec{\phi}$ is distributed as a function of the sample size $N$ and the estimate $\vec{\mu}$. If the prior distribution of $\vec{\phi}$ is given by a Dirichlet distribution then, because the Dirichlet is conjugate with respect to sampling from a multinomial distribution, the posterior distribution of $\vec{\phi}$ after sampling is also a Dirichlet distribution (e.g., see [11]).

## 9   DIRICHLET STOPPING RULES

The distribution $\vec{\mu}$ is computed from the outcomes of $N$ instantiations generated by logic sampling. For example, $\mu_i$ is the fraction of outcomes which instantiate the nodes $\Im$ to $I_i$. The Law of Large Numbers guarantees that in the limit of infinite $N$, $\vec{\mu}$ converges to $\vec{\phi}$ — or equivalently, to $\Pr[\Im]$. For finite $N$, $\vec{\mu}$ is only an approximation of the distribution of $\vec{\phi}$. We consider $\vec{\mu}$ to be a *satisfactory* approximation of $\vec{\phi}$ if, for all $i$, $\mu_i$ and $\phi_i$ satisfy Equation 1. We use the Dirichlet distribution to establish a stopping rule for the number of outcomes $N$ required for $\vec{\mu}$ to be a satisfactory approximation of $\vec{\phi}$.

Given $\vec{\phi}$, the distribution of $\vec{\mu}$ after observing $N$ outcomes of $\zeta$ is given by multinomial distribution. However, since we do not know $\vec{\phi}$ and we do know $\vec{\mu}$, we would like to have a distribution for $\vec{\phi}$ given $\vec{\mu}$ after observing $N$ outcomes. We assume that the distribution of $\vec{\phi}$ prior to making any observations on the outcome of $\zeta$ has a Dirichlet distribution—we consider the implications of this assumption in Section 10. Then, by the conjugate nature of the Dirichlet distribution, the distribution of $\vec{\phi}$ after observing an outcome of $\zeta$ is also a Dirichlet distribution. In particular, using the unbiased-Dirichlet prior $\delta(\vec{\phi}|\vec{0},0)$ to represent the prior distribution on $\vec{\phi}$, $\vec{\phi}$ has distribution $\delta(\vec{\phi}|\vec{\mu}, N)$ after $N$ experiments.

Equation 1 is equivalent to

$$\Pr[\vec{\mu}(1+\epsilon)^{-1} \leq \vec{\phi} \leq \vec{\mu}(1+\epsilon)] > 1 - \delta. \quad (7)$$

Because $\vec{\phi}$ is described by a Dirichlet distribution, the probability term in Equation 7 is given by the cumulative mass of $\delta(\vec{\phi}|\vec{\mu}, N)$ that lies inside the convex polytope defined by the following set of equations:

$$\vec{\phi} \geq \vec{\mu}(1+\epsilon')^{-1} \quad (8)$$
$$\vec{\phi} \leq \vec{\mu}(1+\epsilon'). \quad (9)$$

Conversely, the failure probability $\delta$ is given by the cumulative mass that lies outside the convex polytope,

$$\int_{0 \leq \vec{\phi} < \vec{\mu}(1+\epsilon')^{-1}} \delta(\vec{\phi}|\vec{\mu}, N) d\vec{\phi}$$

$$+ \int_{1 \geq \vec{\phi} > \vec{\mu}(1+\epsilon')} \delta(\vec{\phi}|\vec{\mu}, N) d\vec{\phi} = \delta. \quad (10)$$

Equation 10 allows us to formulate a general probabilistic stopping rule for stochastic simulation algorithms. To achieve an estimate $\vec{\mu}$ of $\vec{\phi}$ that satisfies Equation 1, the stochastic simulation algorithm stops when the left side of Equation 10, evaluated at the current $N$ and $\vec{\mu}$, is less than or equal to the input $\delta$. Details of this analysis can be found in [8].

## 10   STRUCTURE AND EFFECTS OF PRIOR PROBABILITIES

Let us consider the knowledge that an agent has about $\vec{\phi}$ prior to observing the outcome. Possible values of $\vec{\phi}$ must lie in the $K$-cube $[0,1]^K$. Before experimentation, an agent might believe that all values in $[0,1]^K$ are equiprobable. Such a prior distribution is given by a uniform distribution, or, equivalently, by $\delta(\vec{\phi}|(\frac{1}{K},...,\frac{1}{K}), K)$. In the discipline of Bayesian statistics, the distribution $\delta(\vec{\phi}|\vec{0}, 0)$ is considered to be the *unbiased prior* (see, e.g., [14, 12]). The unbiased-Dirichlet prior effectively partitions its mass equally at the vertices of the $K$-cube, reflecting complete uncertainty in $\vec{\phi}$.

Analyses of a preferred prior distribution are rendered immaterial by noting the general insensitivity of results to these alternative prior distributions. The information necessary to update an agent's prior distribution on $\vec{\phi}$ from complete uncertainty—that is, $\delta(\vec{\phi}|\vec{0}, 0)$—to the uniform distribution—that is, $\delta(\vec{\phi}|(\frac{1}{K},...,\frac{1}{K}), K)$—is provided by the first $K$ outcomes. Thus, for large samples, the rate of convergence of the estimate to the mean is insensitive to the choice of an informationless prior distribution on $\vec{\phi}$.

## 11   ANALYSIS OF REFORMULATION TRADEOFFS

Without reformulation of the inference problem $\Pr[X|E]$, BNRAS-algorithms have a runtime proportional to $\mathcal{D}_E^4$ [6, 9]. Reformulation requires us to independently approximate $\Pr[X, E]$ and $\Pr[E]$. The time required to approximate these inferences is proportional to

$$2^{|\Im|}\mathcal{D}_\Im^4 + N. \quad (11)$$

The first term in Equation 11 is the time required to approximate the $2^\Im$ inferences $\Pr[X, E|\Im]$ of Equation 5. The second term is the number of instantiations dictated by the stopping rule that guarantees $\vec{\mu}$ is a satisfactory approximation of $\vec{\phi} = \Pr[\Im]$.

Let $\phi_M$ denote the minimum probability $\Pr[I_i]$ over all instantiations $I_i$. Using results in [7] it is straightfor-

ward to show that

$$N < \frac{2^\Im}{\epsilon^2 \phi_M} \log \frac{2}{\delta}. \quad (12)$$

Equations 11 and 12 imply that the time required for approximation is proportional to

$$2^\Im \mathcal{D}_\Im^4 + 2^\Im \phi_M^{-1}. \quad (13)$$

The choice of nodes $\Im$ that minimizes the time given in Equation 13 requires searching over the space of all possible sets $\Im$. However, even suboptimal selections can be useful; inference based on suboptimal reformulations can be significantly faster than inference on the original problem instance.

We outline a greedy search algorithm that highlights the optimization constraints present in Equation 13. Initially, $\Im$ is empty and the first term in Equation 13 dominates the runtime. Assume that for some nonempty set $\Im$, the first term in Equation 13 continues to dominate. We add $\mathbf{u}_{X_i}$ to $\Im$ only if

$$2^{|\mathbf{u}_{X_i}|} < \lambda_i^\Im, \quad (14)$$

and if, for all $j \neq i$,

$$\frac{\lambda_i^\Im}{2^{|\mathbf{u}_{X_i}|}} > \frac{\lambda_j^\Im}{2^{|\mathbf{u}_{X_j}|}}. \quad (15)$$

Equation 14 guarantees that, if we add $\mathbf{u}_{X_i}$ to $\Im$, the computation time, given by the term $2^\Im \mathcal{D}_\Im^4$ in Equation 13, is decreased. Equation 15 guarantees that adding $\mathbf{u}_{X_i}$ achieves the best reduction of the computation time. The overall computation time in Equation 13 is also decreased because, by assumption, the first term in Equation 13 dominates the running time. The behavior of $\mathcal{D}_\Im^4$ and $\phi_M^{-1}$ is complementary in the sense that, augmenting $\Im$ decreases $\mathcal{D}_\Im^4$ and increases $\phi_M^{-1}$. Thus, a minimal runtime is reached when the two terms in Equation 13 are of comparable magnitude.

## 12 SUMMARY AND CONCLUSIONS

Cooper [5] shows that exact computation of inference probabilities is NP-hard. Thus, for large belief networks, probabilistic inference is intractable if exact results are required. It is equally surprising that the approximation of probabilistic inference is NP-hard. Dagum and Luby [10] show that even crude approximations of inference probabilities can be intractable in certain contexts. Such complexity analyses are sobering with regard to our inability to avoid worst-case intractability. However, the worst-case intractability of exact and approximate inference does not invalidate research on techniques for minimizing inference runtime. Although we cannot avoid worst-case intractability, we can apply methods to refine an initial problem instance by removing unnecessary complexity.

We have described a means of decomposing belief networks by selectively reformulating topologies and conditional probabilities which pose difficult challenges to a stochastic simulation algorithm. Perhaps the most significant aspect of our method is the recruitment of a parameter, developed in a formal analysis of the runtime of an inference approximation algorithm, to serve as an intelligent sentry in targeting the most difficult components of a problem instance for decomposition.

There are opportunities for extending selective conditioning for use in simulation-based inference. For example, we can introduce additional flexibility by integrating bounded conditioning [17] with selective conditioning. With bounded conditioning, we focus the attention of a system on the solution of the most relevant set of subproblems, and consider additional subproblems as time allows. Beyond refining the details of our work with stochastic-simulation-based inference, the general approach of identifying and reformulating troublesome regions of a problem instance holds promise for solving other inference problems, and, perhaps, for tackling difficult computational problems beyond inference.

## Acknowledgments

This work was supported in part by the National Science Foundation under Grant IRI-9108385 and Rockwell International Science Center IR&D funds.